\documentclass[letterpaper]{article} %
\usepackage[]{arxiv23}  %
\usepackage{times}  %
\usepackage{helvet}  %
\usepackage{courier}  %
\usepackage[hyphens]{url}  %
\usepackage{graphicx} %
\usepackage{natbib}  %
\usepackage{caption} %
\usepackage{algorithm}
\usepackage{algorithmic}

\usepackage{newfloat}
\usepackage{listings}
\DeclareCaptionStyle{ruled}{labelfont=normalfont,labelsep=colon,strut=off} %
\floatstyle{ruled}
\newfloat{listing}{tb}{lst}{}
\floatname{listing}{Listing}
\newcommand{\ours}{PRAG}
\newcommand{\optimus}{Optimus}
\newcommand{\nrt}{NRT}
\newcommand{\atttoseq}{Att2Seq}
\newcommand{\peter}{PETER}
\newcommand{\pepler}{PEPLER}

\usepackage{amsfonts}
\usepackage{amsmath}

\title{Factual and Informative Review Generation\\for Explainable Recommendation}

\def\authorspace{\hspace{4mm}}
\author{
    Zhouhang Xie\authorspace{}
    Julian McAuley\authorspace{}
    Bodhisattwa Prasad Majumder\authorspace{}
        \\
        University of California, San Diego \\
        \{zhx022, bmajumde, jmcauley\}\texttt{@ucsd.edu}
        }

\author {
    Zhouhang Xie,\textsuperscript{\rm 1}
    Sameer Singh,\textsuperscript{\rm 2}
    Julian McAuley, \textsuperscript{\rm 1}
    Bodhisattwa Prasad Majumder \textsuperscript{\rm 1}
}
\affiliations {
    \textsuperscript{\rm 1} University of California, San Diego\\
    \textsuperscript{\rm 2} University of California, Irvine\\
    \{zhx022, jmcauley, bmajumde\}@ucsd.edu, sameer@uci.edu
}

\usepackage{bibentry}

\usepackage{booktabs}
\usepackage{multirow}

\usepackage{xcolor}
\definecolor{mypink1}{rgb}{0.858, 0.188, 0.478}
\definecolor{darkred}{rgb}{0.74,0.03,0}
\definecolor{mustardyellow}{rgb}{0.88,0.67,0.01}
\definecolor{navy}{rgb}{0,0,0.5}
\definecolor{darkcyan}{rgb}{0,0.54,0.54}
\definecolor{tabhighlight}{rgb}{0,0.54,0.54}

\newif\ifcomments
\commentstrue
\ifcomments
    \providecommand{\zhouhang}[2][]{{\protect\color{mypink1}{[Zhouhang:\textbf{#1} #2]}}}
    \providecommand{\bodhi}[2][]{{\protect\color{darkcyan}{[Bodhi:\textbf{#1} #2]}}}
    \providecommand{\julian}[2][]{{\protect\color{mustardyellow}{[Julian:\textbf{#1} #2]}}}
    \providecommand{\sameer}[2][]{{\protect\color{pink!50!black}{[Sameer:\textbf{#1} #2]}}}
\else
    \providecommand{\zhouhang}[2][]{}
    \providecommand{\bodhi}[2][]{}
    \providecommand{\julian}[2][]{}
    \providecommand{\sameer}[2][]{}
\fi

\begin{document}

\maketitle

\begin{abstract}
Recent models can generate fluent and grammatical synthetic reviews while accurately predicting user ratings. 
The generated reviews, expressing users' estimated opinions towards related products, are often viewed as natural language ‘rationales’ for the jointly predicted rating. 
However, previous studies found that existing models often generate repetitive, universally applicable, and generic explanations, resulting in uninformative rationales.
Further, our analysis shows that previous models' generated content often contain factual hallucinations.
These issues call for novel solutions that could generate both \textit{informative} and \emph{factually grounded} explanations.
Inspired by recent success in using retrieved content in addition to parametric knowledge for generation, we propose to augment the generator with a personalized retriever, where the retriever's output serves as external knowledge for enhancing the generator.
Experiments on Yelp, TripAdvisor, and Amazon Movie Reviews dataset show our model could generate explanations that more reliably entail existing reviews, are more diverse, and are rated more informative by human evaluators.
\end{abstract}

\section{Introduction}

Recently, there has been increasing interest in treating review generation as a proxy for explainable recommendation, where generated reviews serve as rationales for the models' recommendations\cite{li-etal-2016-diversity, li-etal-2021-personalized, ni-etal-2017-estimating, ni-mcauley-2018-personalized}.
However, existing models commonly generate repetitive and generic content, resulting in uninformative explanations~\cite{geng-etal-2022-improving}.
Further, when evaluating the factuality of generated reviews using pre-trained entailment model, our analysis shows 
that existing models are also susceptible to factual hallucination, a long-existing challenge in many natural language generation (NLG) tasks~\cite{pagnoni-etal-2021-understanding, maynez-etal-2020-faithfulness}.
Specifically, the models often generate statements that are not supported by information about the corresponding product in the training set.
Both nonfactual and uninformative explanations are undesirable, as end users would look for recommendation rationales that truthfully reflect the characteristics of the product without being overly generic.
Thus, these problems limit the usability of natural language explanations (NLE) produced by existing explainable recommendation models.

In order to address the issue that models commonly generate univiersally correct explanations, previous works experimented with diversifying generated reviews using distantly retrieved images as additional signals~\cite{geng-etal-2022-improving}.
However, recommender system datasets do not always have associated images, and \citeauthor{geng-etal-2022-improving} proposed to retrieve images from the web using available textual data.
While this method indeed significantly diversifies the generated natural language explanations (NLE), there is no guarantee that the retrieved content will truthfully represent the quality of the corresponding product.
Thus, the generator needs to condition on a given feature or aspect the user cares about at inference time, commonly extracted from the \textit{ground-truth} review.
This limits the usability of models as such user input might not be available at inference time. 
Another line of work attempted to incorporate a pre-trained language model for better generation quality~\cite{li-etal-2021-personalized}. 
However, the same study shows that pre-trained language models such as GPT-2 struggle to produce diverse reviews while maintaining competitive recommendation accuracy. 
Thus, generating informative and factual reviews without having access to information in ground truth reviews remains an open problem.

\begin{figure*}[tb]
\centering
\includegraphics[trim= 30 375 90 250,clip,width=0.95\textwidth]{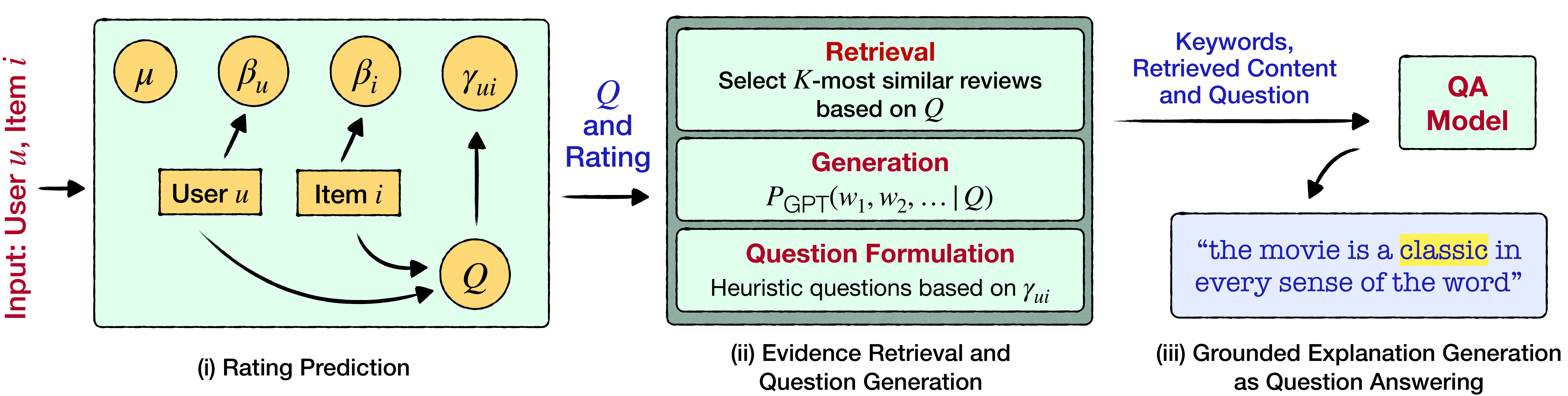}
\caption{\textbf{The proposed framework \ours}. Given the input user and item, our model produces a rating estimation (Section~\ref{sec:rating__prediction}), as well as a latent query $Q$ (Section~\ref{subsec:review_embedding_prediction}). We then use the query to retrieve existing reviews as evidence for the recommendation, and probe a question answering model for natural language explanation (Section~\ref{sec:qa_explanation_generation}). 
}
\label{fig:prag_architecture}
\end{figure*}

Recent advances in knowledge-grounded NLG show that retrieving unstructured text as supplements for the model’s parametric knowledge significantly improves factuality and diversity of generated content~\cite{DBLP:conf/nips/LewisPPPKGKLYR020, guu2020realm}.
Inspired by such success, in this work, we propose to leverage existing reviews as additional context for generating recommendation explanations.
Specifically, we propose \textbf{P}ersonalized \textbf{R}etriever \textbf{A}ugmented \textbf{G}enerator (\textbf{\ours}), a model loosely based on the retriever-reader framework for explainable recommendation.
\ours~consists of a personalized retriever that 
can
accurately give a rating estimation and generate a latent query for review prediction given the input user and item.
To encourage \textit{factual} explanations, we formulate the task of NLE generation as question-answering, where a reader model produces the explanation grounded on retrieved content. 
Meanwhile, to ensure the \textit{informativeness} of the explanations, we estimate a set of personalized high-tf-idf keywords using the latent query, and use these keywords to guide the answer of the reader model. In this way, the reader model produces the final explanation by abstracting input text snippets and keywords, which yields superior quality compared to previous work in both automatic and human evaluation. Our contributions are as follows: 
\begin{itemize}
\item To our best knowledge, we are the first work that evaluates and highlights the importance of factuality in natural language explanation for recommender systems.
\item We develop a novel personalized retriever-reader model for generating factual and informative recommendation explanations.
\item We personalize question-answering for generating recommendation explanations.
\end{itemize}

\section{Related Work}

Explainable recommendation aims at providing users insight for a recommender systems' decision. 
Following earlier works that provide topic words and product features as explanations~\cite{mcauley2013hidden, DBLP:conf/sigir/ZhangL0ZLM14}, recent works are increasingly focusing on generating reviews as explanations~\cite{ni-etal-2017-estimating,ni-mcauley-2018-personalized, Hada2021ReXPlugER, dong-etal-2017-learning-generate}. 
However, these works 
are not focused on the factuality of the generated content. 
Meanwhile, existing works require training a language model from scratch, while previous attempts at leveraging pre-trained language models either require decoding-time search~\cite{Hada2021ReXPlugER}, or are not as performant as other recommendation models~\cite{2022-PEPLER}.

It is to be noted that there are commonly used existing metrics such as Distinct-N~\cite{DBLP:conf/acl/LiuSZK0H22} and unique sentence ratio (USR)~\cite{li_2020_generate_neural_template} that focus on diversity of generated explanations. 
However, a common measure used to increase diversity is to generate a sentence based on words and phrases from the \textit{ground-truth} review, such as in~\citeauthor{geng-etal-2022-improving} and \citeauthor{ni-mcauley-2018-personalized}. 
Our work differs from previous works in that we consider specifically the case where no information from the ground truth review is given, which is common in recommender systems. 

Outside of recommender systems, there is a general trend for using natural language as explanations for various tasks, such as  text classification~\cite{hancock-etal-2018-training, DBLP:conf/nips/ZhouHZLSXT20}, image captioning~\cite{DBLP:journals/corr/abs-2106-13876, marasovic-etal-2020-natural} and question answering~\cite{dalvi-etal-2021-explaining, DBLP:journals/corr/macaw}. However, expert annotated explanations are usually unavailable in recommender system datasets, and reviews are usually noisy by nature and require further processing ~\cite{brazinskas-etal-2020-unsupervised, brazinskas-etal-2021-learning}. This leaves learning to generate explanations from noisy supervision an open challenge, which we seek to address in this work.

Another related
existing problem in natural language generation tasks is safe and generic outputs. 
For example, this problem is well studied in dialogue generation, where previous work shows that exposure bias and maximum likelihood objective lead the model to produce universally applicable responses~\cite{zhou-etal-2021-learning}. 
In another parallel line of work on evaluating natural language explanations, \citeauthor{wiegreffe-etal-2021-measuring} pointed out that universally correct explanations are undesirable, calling for diverse and informative explanations. 
However, these works commonly focus on traditional NLP tasks and well-formulated datasets.

\section{\ours: Setup and Overview}

We first introduce the joint review-rating prediction task, then cover our model's general architecture.

\subsection{Problem Setup}

Consider a set of users $U$ and items $I$, where each user $u \in U$ and item $i \in I$ in the training set is associated with a real numbered rating $r_{u, i} \in \mathcal{T}$ and a review $e_{u,i} \in E$ that serves as explanations for the users rating decision. The task of joint rating-explanation estimation learns a function $\operatorname{rec}: u, i \to \hat{r}_{u,i}, \hat{e}_{u,i}$, where $\hat{e}_{u,i}$ is a textual explanation that informs the user of the reasoning behind $\hat{r}_{u,i}$. Note that broadly speaking, explanations can be in various forms, such as topic words and feature importance, here, we use $e_{u,i}$ to denote natural language explanation specifically.

\subsection{Model Overview}

The architecture of \ours~is depicted in Figure~\ref{fig:prag_architecture}. 
Given the input user $u$ and item $i$, we first obtain semantic embeddings for all related existing reviews.
Here, ``related reviews" are reviews written by the user or written about the item.
Then, given the historical reviews, a personalized retriever model produces a latent query $Q_{u,i}$ that is close to the ground truth explanation given the input user and item. 
A rating prediction module then uses such a latent query to produce a rating estimation.

To generate an explanation, we exploit the rich semantic information in $Q_{u,i}$.
Specifically, we (1) retrieve a set of existing reviews $G$ based on $Q_{u,i}$, where $G \subseteq E$, and (2) generate a set of high tf-idf scored keywords that the user would care about based on $Q_{u,i}$ and $G$.
Finally, a keyword-guided question answering model produces the final explanation based on the keywords, retrieved reviews, and the question.
We now provide descriptions for each component.

\section{Personalized Retriever}

\begin{figure}[tb]
\centering
\includegraphics[trim= 200 320 890 120,clip,width=0.9\linewidth]{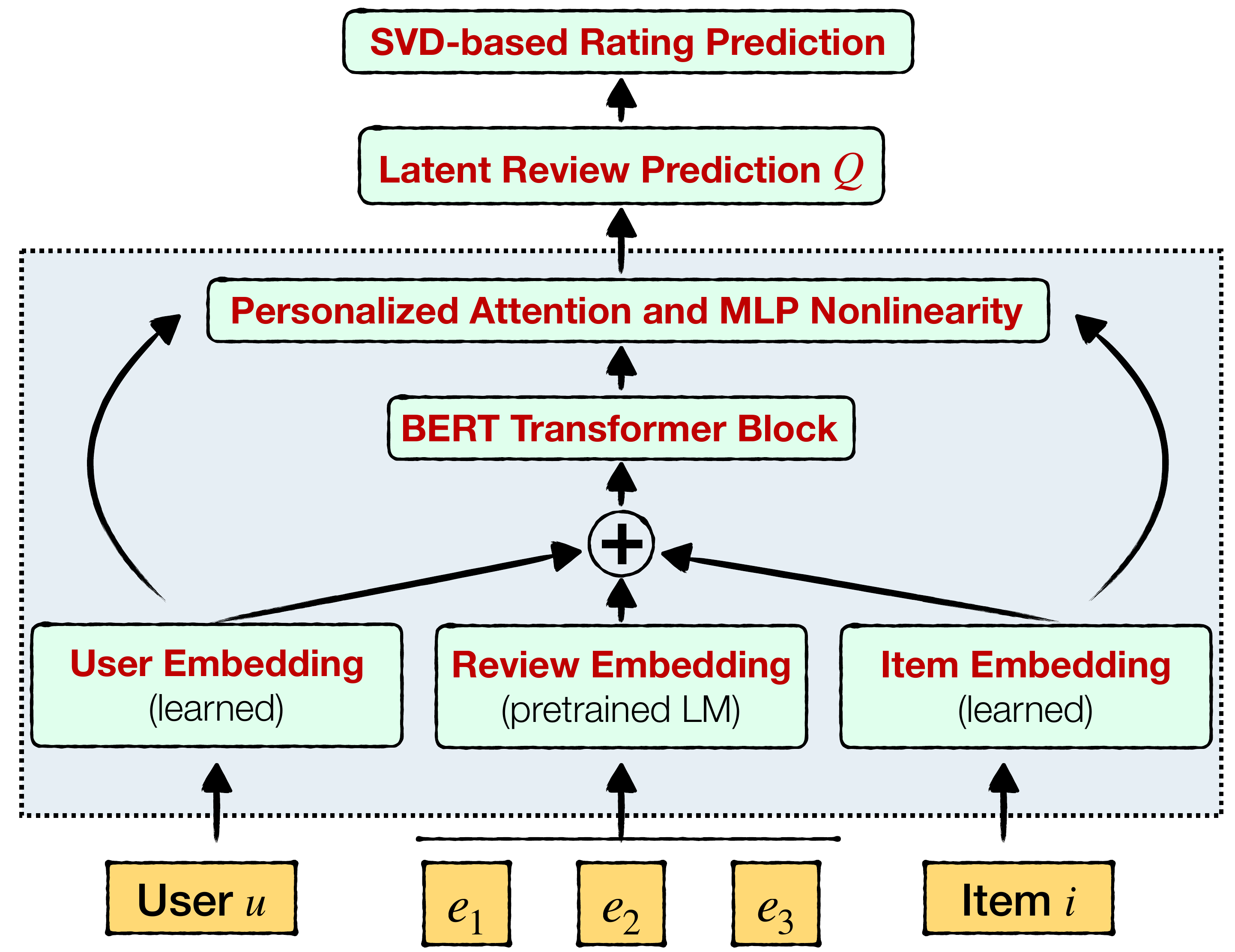}
\caption{Overview for the \textbf{retriever architecture}. Our model takes as input user and item id and a set of corresponding historical reviews, then produces a latent review $Q$. The latent review is then used for rating prediction, retrieval, and explanation generation.}
\label{fig:retriever_arch}
\end{figure}

The architecture of our retriever model is as shown in Figure~\ref{fig:retriever_arch}. 
The review, item, and user embedding go through a stack of transformer blocks and are then pooled to generate the estimated review embedding. 

\subsection{Embedding Reviews}

Pre-trained language model representations contain rich semantic information about sentences and supports arithmetic operations such as addition and dot product. 
Following previous success in pre-computing embeddings for efficient retrieval, such as in RAG \cite{DBLP:conf/nips/LewisPPPKGKLYR020}, we obtain sentence embeddings for all reviews using a pre-trained sentence encoder $\mathrm{\Phi}$ that outputs $d$ dimensional sentence representations. 
In practice, we adopt the MPNET~\cite{DBLP:conf/nips/Song0QLL20} model from huggingface transformers~\cite{wolf-etal-2020-transformers} for balanced efficiency and embedding quality. 
We note any state-of-the-art general-purpose sentence encoding model can be used for \ours, and provide additional analysis with T5~\cite{DBLP:journals/jmlr/RaffelSRLNMZLL20} sentence encoders in our experiment section.

\subsection{Review Aggregation}

Our base personalized retriever model is based on BERT architecture~\cite{devlin-etal-2019-bert}. 
Specifically, we treat each historical review in the training corpus as a token. When predicting the review a user would give to an item, the input to the model is a set of historical reviews written by the user and written for the item. 
Since these tokens do not represent consecutive words in a sentence as in the original BERT model, we do not use any position embedding. 
However, historical reviews play different roles when they are related to the item versus being related to the user, thus we choose to maintain learnable embedding $c \in \mathbb{R}^{d}$ to represent these two distinct scenarios, which are added to the original review embedding instead of position embeddings. 

Finally, we maintain a learnable embedding $v_{u} \in \mathbb{R}^{d}$ and $v_{i} \in \mathbb{R}^{d}$ for each user and item to model user preferences, following common practice in recommender systems. 
For the input user and item, we look up the corresponding embedding and add these embeddings to each of the input review embedding. 
The final input review embedding to the model is thus:
 \[
    v_{u}+v_{i}+c.
\]

Such a final embedding is then passed through a stack of 2-layer transformers to process the cross-review relationship in the input.

\subsection{Personalized Attention}

Not every review in the input is important. 
Thus, it is crucial to select reviews that are helpful to the recommendation model. 
To achieve this, we develop a personalized attention module for weighing review representations from the base transformer model. 
Specifically, we obtain an attention score with respect to each piece of input reviews using a standard linear layer with a Rectified Linear Unit (ReLU) activation function using the concatenation of the review embedding, user embedding, and item embedding. 
Following previous works' insight that recommendation models benefit from un-smoothed attention scores' ability to discard irrelevant items~\cite{Zhou2018DIN}, we conduct weighted pooling directly using the normalized attention scores.

\subsection{Review Embedding Prediction}
\label{subsec:review_embedding_prediction}

Finally, the weighted sum of review embeddings is passed through a multi-layer-perception (MLP) layer to produce the final latent query $Q_{u,i} \in \mathbb{R}^{d}$. 
During training, we minimize the $\mathrm{L2}$ distance between the produced query and the embedding of the corresponding ground truth review, $\mathcal{L}_\text{retrieve}$, following previous success in using vectorized reviews to regularize recommender models~\cite{Hada2021ReXPlugER}. 
Such embedding thus represents the predicted semantics of the input user and item, which could then be used to retrieve relevant reviews that are semantically similar to the predicted review from the existing review corpus.

\subsection{Rating Prediction}
\label{sec:rating__prediction}

To perform rating prediction, we combine HFT~\cite{mcauley2013hidden}, a strong matrix-factorization based explainable recommendation model with a modified wide-and-deep~\cite{cheng2016widedeep} architecture. Specifically, the original HFT model makes predictions by modifying the following equation:
\[
    \operatorname{rec}(u,i) = \gamma_u \times \gamma_i + \beta_u + \beta_i + \mu,
\]
where $\mu$ is set to the global mean value of all ratings, and $\beta_{u}, \beta_{i}$ are the learned bias. 
The model further ties either $\gamma_u$ or $\gamma_u$ to topic models, and learns product or user-specific topics by jointly minimizing the rating regression loss and the negative likelihood of the corresponding review. 

We extend the HFT model by using latent query $Q_{u,i}$.
Specifically, we adopt the estimated latent review as a new source of semantic information in place of topic models. 
For example, when associating item features to semantic information, we use a simple multi-layer-perception (MLP) to map $Q_{u,i}$ to $\gamma_i$ (or $\gamma_u$). 
Meanwhile, previous work has shown that using a shallow linear (wide) layer for memorizing simple patterns could increase the performance of recommendation models~\cite{cheng2016widedeep}. 
Following this intuition, we add an additional linear (wide) layer to the original model
\[
    \operatorname{rec}(u,i) = \operatorname{MLP}(Q) \times \gamma_i + \operatorname{wide}(Q) + \beta_u + \beta_i + \mu.
\]

We learn to predict rating using the standard squared loss $\mathcal{L}_\text{rating}$.
The final joint loss for training the personalized retriever is then
\[
    \mathcal{L}_\text{retrieve} + \mathcal{L}_\text{rating}.
\]

\section{Explanation Generation as Keyword-guided Question Answering}
\label{sec:qa_explanation_generation}

\begin{figure}[t!]
\centering
\includegraphics[trim= 490 230 650 230,clip,width=0.75\linewidth]{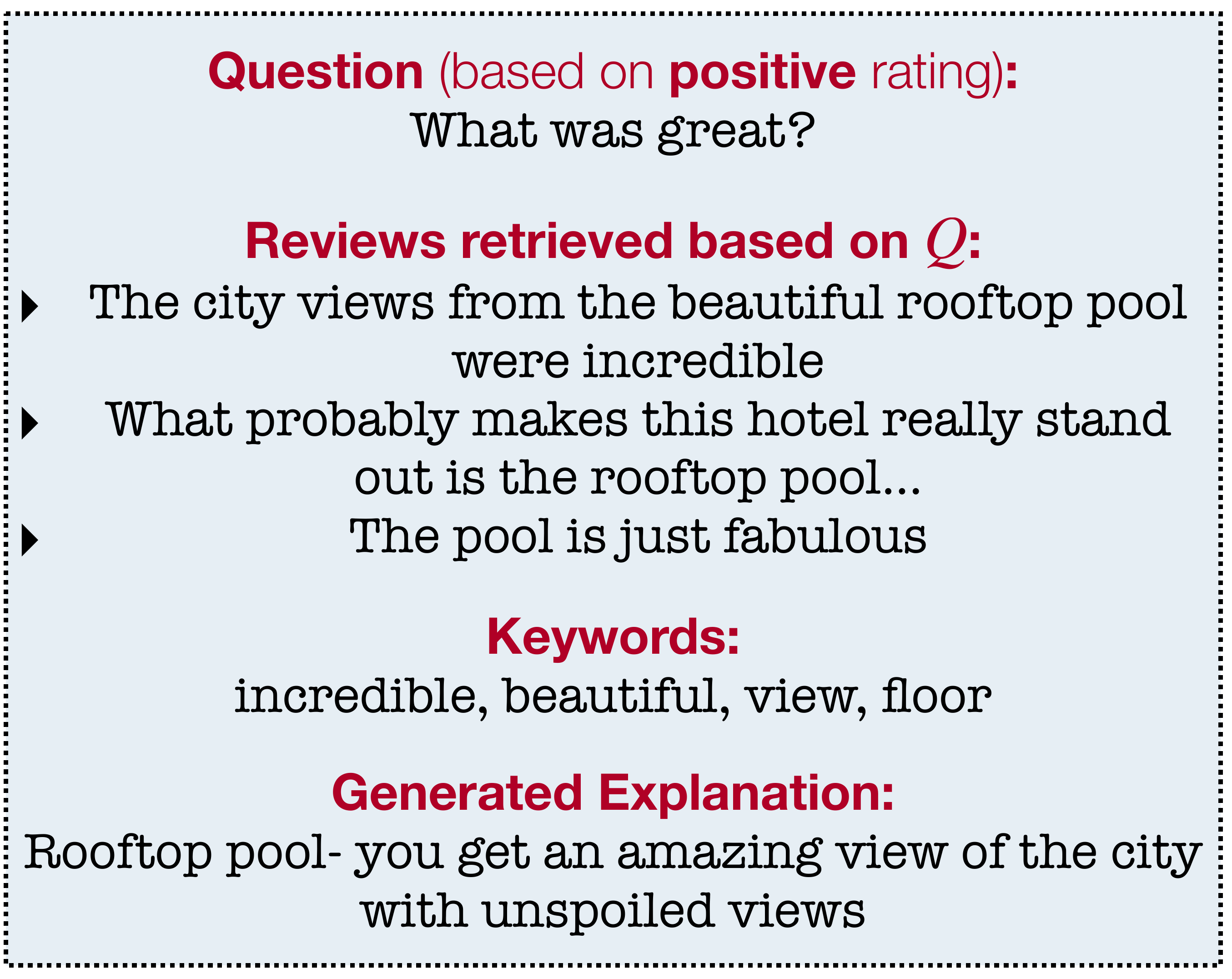}
\caption{
\textbf{\ours}~use retrieved historical reviews as source text, and \textbf{generates} the explanation based on the keywords.}
\label{fig:explanation_generation}
\end{figure}

To generate an explanation, we source information from retrieved reviews and treat the task of explanation generation as question answering.
Specifically, the reader model should be able to answer why a higher (or lower) rating \textit{adjustment score} $\gamma_{ui}$ is being produced, as this is the only inter-user-item factor in rating prediction.
To achieve this, we first train an embedding estimator that generates \textit{informative} keywords using the mean embedding of latent query $Q_{u,i}$ as well as its corresponding set of retrieved reviews.
After this, we probe a question-answering model trained to \textit{factually} reflect content from the input with natural language prompts while incorporating the generated keywords.
The schema of the explanation generation pipeline is as shown in Figure~\ref{fig:explanation_generation}.

To facilitate factual behavior for both the embedding estimator and the question-answering model, we design an aggregated task where the ground-truth is \textit{guaranteed} to have a strong correlation to model input at training time. 
Concretely, we train a model to recover informative keywords in a review from the latent vector encoded from that specific review.
Now we discuss each of the components in detail.

\subsection{Retrieving Reviews}
Given that the latent query is optimized to be similar to the ground truth review, a natural scheme for review retrieval is to rank existing reviews' semantic embedding with respect to $Q_{u,i}$ using similarity metrics. 
However, in practice, we found the model often produces overly-generic retrieval results, i.e., the reviews for multiple users tend to be similar, as observed in previous work~\cite{geng-etal-2022-improving}.
To address this issue, we propose to retrieve existing reviews using characteristics that are \textit{specific} to the user.
In other words, we want to de-emphasize explanations that a model will produce for \textit{every} user based on the item.
We do so by examining the explanation for other users, estimating the latent explanation that are produced for all users, and marginalizing out such a universal explanation.

Concretely, we sample a batch of users at inference time, and obtain each sampled user's corresponding latent query $Q_{{u_n},i}$. 
Then, we estimate the explanation that would be produced for every user using the mean embedding of the batch of predicted queries.
Finally, we subtract such mean embedding from the original $Q_{u,i}$, effectively marginalizing out the user-agnostic aspect of the latent query. 
Since this changes the magnitude of $Q_{u,i}$, we rank existing reviews using cosine distance instead of $\mathrm{L2}$ distance, and select top reviews as the retrieved content. 
Note that if $Q_{u,i}$ is tied to $\gamma_i$, we marginalize out the universal explanation for a batch of items instead, as we need to de-emphasize the explanation being produced for every item.
We provide qualitative and quantitative analysis of such marginalization in Section~\ref{sec:verifying_decision_choices}. 

\subsection{Informative Keyword Generation}

Retrieved reviews are by nature noisy, and multiple aspects in the reviews could simultaneously contribute to the model's rating decision.
However, to generate realistic reviews, the generator must select a few aspects to focus on, as natural reviews typically focus on a few central topic words.
To ensure the informativeness of the generated review, we design an embedding estimator that generates high tf-idf scored keywords given the personalized latent query.
We then use the keywords in the explanation generation phase for improved informativeness.

To achieve this, we adopt a pre-trained GPT-2 model\footnote{\url{https://huggingface.co/distilgpt2}}, and formulate the optimization objective as the task of embedding-to-keyword prediction. 
Specifically, given a latent embedding $\Phi(e)$ encoded from an original review $e$ in the training set, we train the embedding estimator on recovering the 5-highest tf-idf scored words from the original review. 
In practice, we use the concatenated target keywords as desired model output and fine-tune the language model using the standard MLE objective.
At inference time, we could then condition the embedding estimator on the corresponding latent query since the query is optimized to be in the semantic space of the sentence encoder $\Phi$.

\subsection{Explanation Generation}

For explanation generation, we adopt a general purpose question-answering model\footnote{\url{https://github.com/allenai/unifiedqa}} 
, and exploit the in-context learning ability of large-scale, pre-trained language models~\cite{Brown2020LanguageMA}. 
Specifically, we ask the model to answer the question ``\texttt{what was great?}" or ``\texttt{what was not good?}", depending on the sign of the predicted rating adjustment score.
Meanwhile, we note that the choice of prompt does not significantly influence the generation quality, and provide experiment with alternative prompts in the appendix.

To guide the model to cover the aspect that the user cares about, we want to encourage the model's output to contain at least one of the keywords produced by the embedding estimator.
To achieve this, we first apply constrained decoding~\cite{hokamp-liu-2017-lexically} using the original, unguided question answering model, forcing the output to include at least one of the input keywords. 
However, this could result in ungrammatical text. 
To address this issue, we manually rephrase 100 outputs for each dataset, which is easily achievable.
This results in a high-quality dataset where each input text is paired with a set of keywords, with a ground-truth answer that contains at least one of the keywords. 
We then fine-tune the original question answering model on this dataset for a maximum of 10 epochs, following previous work on few-shot question answering~\cite{ram-etal-2021-shot}. 
The fine-tuned model could then be applied for efficient explanation generation \textit{without} any decoding-time constraints.

\section{Experiments}

\subsection{Datasets}

We conduct our experiments on three publicly available dataset and splits from various domains~\cite{li_2020_generate_neural_template}: Yelp\footnote{https://www.yelp.com/dataset/challenge} (restaurant reviews), TripAdvisor\footnote{https://www.tripadvisor.com
} (hotel) and Amazon Movies and TV~\cite{He2016UpsAD}.
Note that the data splits guarantee that products in the test set always appear in the training set.

\begin{table*}[h]
\small
\centering
\setlength\tabcolsep{4pt}
\resizebox{\textwidth}{!}{%
\begin{tabular}{l cc cc cc c cc cc cc c cc cc cc c cc cc cc c cc cc cc c cc cc cc c}
\toprule
\multirow{2}{*}{}
  & \multicolumn{3}{c}{\textbf{Entail}}
  &
  & \multicolumn{3}{c}{\textbf{D-1}} 
  &
  & \multicolumn{3}{c}{\textbf{D-2}}
  &
  & \multicolumn{3}{c}{\textbf{ENTR}}
  &
  & \multicolumn{3}{c}{\textbf{USR}}
  &
  & \multicolumn{3}{c}{\textbf{MAUVE}}\\
\cmidrule(lr){2-5}
\cmidrule(lr){6-9}
\cmidrule(lr){10-13}
\cmidrule(lr){14-17}
\cmidrule(lr){18-21}
\cmidrule(lr){22-25}
  \textbf{Method}
  & 
  \multicolumn{1}{c}{\textbf{Movie}} & \multicolumn{1}{c}{\textbf{Trip}} & \multicolumn{1}{c}{\textbf{Yelp}} &
  &
  \multicolumn{1}{c}{\textbf{Movie}} & \multicolumn{1}{c}{\textbf{Trip}} & \multicolumn{1}{c}{\textbf{Yelp}} &
  &
  \multicolumn{1}{c}{\textbf{Movie}} & \multicolumn{1}{c}{\textbf{Trip}} & \multicolumn{1}{c}{\textbf{Yelp}} &
  & 
  \multicolumn{1}{c}{\textbf{Movie}} & \multicolumn{1}{c}{\textbf{Trip}} & \multicolumn{1}{c}{\textbf{Yelp}} &
  &
  \multicolumn{1}{c}{\textbf{Movie}} & \multicolumn{1}{c}{\textbf{Trip}} & \multicolumn{1}{c}{\textbf{Yelp}} &
  &
  \multicolumn{1}{c}{\textbf{Movie}} & \multicolumn{1}{c}{\textbf{Trip}} & \multicolumn{1}{c}{\textbf{Yelp}} &
\\
\midrule
Att2Seq
               & 25.6
               & 12.2
               & 35.9 
               &
               & 39.9
               & 34.6        
               & 43.1  
               & 
               & 75.9
               & 75.4         
               & 78.1 
               & 
               & 9.56
               & 8.11
               & 8.44
               &
               & 41.7
               & 21.0
               & 39.9
               &
               & 3.0
               & 1.4
               & 3.9
\\
NRT
               & 36.1
               & 10.0
               & 31.4 
               &
               & \underline{44.0}
               & 32.4          
               & 41.0 
               & 
               & 77.8
               & 72.8          
               & 76.6  
               & 
               & 7.5
               & 7.5
               & 8.3
               &
               & 36.1
               & 46.3
               & 44.4
               &
               & 3.0
               & 3.0
               & 4.2
\\
\peter
               & 29.0
               & 17.5
               & 44.5 
               &
               & 27.7
               & 26.8          
               & 29.5
               & 
               & 58.6
               & 60.7          
               & 60.4  
               & 
               & 10.5
               & 10.1
               & 10.7
               &
               & 60.7
               & 57.2
               & 58.2
               &
               & 3.7
               & 2.3
               & 2.2
\\
PEPLER
               & 17.9
               & 11.0
               & 16.0
               &
               & 23.2
               & 23          
               & 25.5
               & 
               & 51.5
               & 52.2          
               & 52.5  
               & 
               & 11.1
               & 10.0
               & 11.0
               &
               & 52.6
               & 41.7
               & 49.1
               &
               & 1.1
               & 0.4
               & 0.4
\\
\midrule
\optimus
               & 25.1
               & 22.8
               & 11.5
               &
               & 31.9
               & 32.8        
               & 33.2
               & 
               & 77.3
               & 77          
               & 79.3  
               &
               & 10.3
               & 8.5
               & 10.7
               &
               & \textbf{98.5}
               & \underline{92.1}
               & \textbf{96.1}
               &
               & 3.5
               & 3.3
               & 4.5
\\
SUM
               & \underline{49}
               & 29.5
               & 30.8
               &
               & 22.1
               & 18.7        
               & 20
               & 
               & 67.1
               & 61          
               & 63.7  
               & 
               & \underline{11.2}
               & \underline{10.4}
               & \underline{11.5}
               &
               & \underline{95.3}
               & \textbf{94.7}
               & \underline{94.8}
               &
               & 5.8
               & 4.7
               & 5.4
\\
PETER+
               & 40.0
               & \underline{32.6}
               & \underline{59.4}
               &
               & 43.9
               & \textbf{42.6}        
               & \underline{47.0}
               & 
               & \underline{78.4}
               & \underline{81.9}        
               & \underline{83.1}  
               & 
               & 9.48
               & 8.53
               & 9.85
               &
               & 60.6
               & 31.5
               & 52.8
               &
               & \underline{12.9}
               & \underline{5.3}
               & \underline{10.4}
\\
PRAG
               & \textbf{88.8}
               & \textbf{80.1}
               & \textbf{86.2}
               &
               & \textbf{45.6}
               & \underline{39.9}       
               & \textbf{47.1}       
               & 
               & \textbf{84.3}
               & \textbf{82.2}        
               & \textbf{84.7}
               & 
               & \textbf{12.0}
               & \textbf{12.0}
               & \textbf{11.9}
               &
               & 71.8
               & 76.5
               & 70.4
               &
               & \textbf{23.1}
               & \textbf{42.8}
               & \textbf{20.3}
\\
    \bottomrule
    \end{tabular}%
}
    \caption{\textbf{Automatic evaluation results} on test sets. \textbf{Entail} denotes the percentage of entailment among the generated reviews. \textbf{D-1} and \textbf{D-2} scores denote Distinct-1 and Distinct-2 scores.
    \textbf{ENTR} denotes the geometric mean of n-gram entropy among the generated text. \textbf{USR} denotes the ratio of unique sentences generated. \textbf{MAUVE} is a distribution based-metric to measure the quality of the generated text. Best and second-best performances are bolded and underlined, respectively. See Section~\ref{sec:automic_evaluation} for details.} 
\label{tab:main_results}
\end{table*}

\subsection{Baselines}

We compare four commonly used models in the literature: \atttoseq~\cite{dong-etal-2017-learning-generate}, \nrt~\cite{Li2017NeuralRR}, \peter~\cite{li-etal-2021-personalized}, and \pepler~\cite{2022-PEPLER}. Among the models, \atttoseq~and \nrt~are LSTM-based models, while \peter~is a transformer-based model.
We additionally incorporate a variant of \peter~that conditions on a topic word from ground-truth review at inference time, denoted by \peter+.
Finally, \pepler~adopts a pre-trained GPT-2 model with prompt tuning for explanation generation.
We provide more details for each of the models in the appendix.

Further, we show that when augmented with our personalized retriever, opinion aggregation approaches such as summarization models could seamlessly integrate into the retriever-reader framework. 
We demonstrate this by proposing two \textit{novel} baselines, where summarization models are used to aggregate retrieved reviews.

\paragraph{\ours-Optimus (\optimus)} 
Adopts a state-of-the-art pre-trained VAE-based language model based on BERT and GPT~\cite{li-etal-2020-optimus}. 
Recent studies find VAE-based language models could be used for unsupervised opinion summarization~\cite{iso-etal-2021-convex-aggregation}. 
However, \citeauthor{iso-etal-2021-convex-aggregation}'s approach requires searching over large amounts of potential sequences at inference time, particularly when there are multiple inputs. 
Thus, we use the original Optimus model without inference-time searching as our baseline.
Specifically, we fine-tune a pre-trained Optimus model on each dataset for 1 epoch on language modeling, per recommendation by the original authors, and condition the GPT-based generator on the mean embedding of retrieved reviews at inference time.

\paragraph{\ours-SUM (SUM)} 
Following previous success in training summarizers by learning to recover the target review from a set of distantly retrieved similar reviews~\cite{amplayo-lapata-2020-unsupervised}, we train our summarizer in a leave-one-out fashion over sets of similar reviews.  
Specifically, for each 
review in the training set, we retrieve a set of most similar reviews using cosine distance in sentence encoder $\Phi$'s semantic space. 
These retrieved reviews then serve as the input for the model. 
Then, the summarization module is trained to recover the original review that is used for retrieval from the retrieved reviews.
For training, we use the same pre-trained T5 model as \ours~as our base model for a fair comparison and train the model using maximum likelihood estimation (MLE) objective. 
We provide more details for data processing and training of SUM in the appendix.

\subsection{Automatic Evaluation}

\label{sec:automic_evaluation}

For automatic evaluation, we generate 10,000 samples from each baseline model, and measure the performance in terms of factuality, informativeness, and generation quality. 
We provide additional recommendation performance analysis on all eligible models on the whole test set.

\paragraph{Factuality.}
Factuality constraints that the models' generated explanations are factually correct. 
Intuitively, the statement in a generated explanation is correct if it could be supported by any existing reviews in the training data. Otherwise, the model is exhibiting \textbf{hallucinations}~\cite{maynez-etal-2020-faithfulness}. 
As reported by the same work, entailment models could better measure the factual consistency between reference text and generated content. 
We follow their findings and evaluate the entailment relationship between generated reviews with reviews of the same product in the training set. 
Specifically, for each generated explanation, we check whether the explanation entails \textit{any} reviews from the training set for the same product using a pre-trained entailment model~\footnote{\url{https://huggingface.co/prajjwal1/roberta-large-mnli}}. 
We note that a generated explanation is factual if it entails any piece of existing reviews for the product, and report the entailment ratio.
Specifically, we report the percentage of entailment as Entail.

\paragraph{Informativeness.}
We measure the informativeness of generated explanations using token, sentence, and corpus level evaluations.
Concretely, We evaluate the models using Distinct-1 and Distinct-2 (D-1, D-2) scores~\cite{li-etal-2016-diversity}, Unique Sentence Ratio (USR) as proposed by~\cite{li_2020_generate_neural_template} and ENTR~\cite{jhamtani-etal-2018-learning} following previous work on diversifying generated content~\cite{majumder-etal-2021-unsupervised}.

\paragraph{Generation quality.}
To measure the generated explanation quality, we opt to measure how human-like the generated explanations are. 
Specifically, we adopt MAUVE~\cite{pillutla-etal:mauve:neurips2021}, a distribution-based evaluation metric that measures how close the generated contents are to the ground truth corpus.

\paragraph{Recommendation performance.}
We evaluate the recommendation performance using Rooted Mean Squared Error (RMSE) score following previous works done using the same dataset.

\section{Results}

\subsection{Quantitative Evaluation}

\label{sec:quantitative_results}

We report PRAG's performance compared to baseline models as in Table~\ref{tab:main_results}. 
\ours~consistently outperforms the baseline models in terms of both diversity and informativeness, and can generate high-quality sentences compared to human-written reviews. 

\paragraph{Retrieval component improves generation quality.}
As shown in Table~\ref{tab:main_results}, conditioning on additional information improves both model diversity and factuality. 
Specifically, \optimus, SUM, and \ours~consistently outperform other models that only maintain vector representation for each user and item. 
The generated content has both higher sentence-wise diversity (from D-1 and D-2 scores) and higher corpus-level diversity (from ENTR and USR scores), as well as being closer to human written reviews (from MAUVE scores). 
This shows that the personalized retrieval component could improve generation quality. 

\paragraph{Training data affects generator factuality.
}
While retrieved historical reviews encourage more factual output in general, this does not guarantee strong factuality. 
In particular, the T5-based summarizer and \optimus~model have limited improvement in terms of factuality compared to other baseline models. 
This highlights the major cause of hallucination is directly training generators on noisy data. 
Specifically, \citeauthor{maynez-etal-2020-faithfulness} reported that ground-truth sequences in the training data that contain hallucinated content would trigger the model to be less factual at inference time in summarization. 
In parallel, \citeauthor{longpre-etal-2021-entity} reported that noisy retrieved content would cause the generator to hallucinate for question answering. 

We note that a similar case applies to review generation as well. 
In particular, there is no guarantee that the ground truth review will entail \textit{any} of other reviews for the same product in the training set. 
While this behavior is natural for human~\cite{maynez-etal-2020-faithfulness}, the model ended up learning to hallucinate during optimization as a result.
By leveraging a reader model trained to faithfully present content in the input text, \ours~sidesteps this issue, and thus has the best factuality performance across models.

\subsection{Recommendation Performance}

\begin{table}[tb]
\small
\setlength\tabcolsep{11pt}
\center
\begin{tabular}{lccccc}
\toprule
{} &  \bf Movie &  \bf TripAdivor &  \bf Yelp  \\
\midrule
\nrt  &  0.79 & 0.95 &      1.01  \\
\peter  & 0.80 & 0.95 &        1.01  \\
\pepler  &  1.71   &  1.25 &    1.69  \\
\ours$_{u}$ & 0.80 & 0.95 & 1.01 \\
\ours$_{T5}$ & 0.80 & 0.96 & 1.00 \\
\ours~&  0.79 & 0.95 &      1.01  \\
\bottomrule
\end{tabular}
\caption{\textbf{RMSE scores for recommendation performance}. By default, \ours~uses MPNET sentence embedding and ties $Q_{u,i}$ to $\gamma_{i}$. \ours~$_{T5}$ uses embedding from T5 model and PRAG(u) maps $Q_{u,i}$ to $\gamma_u$.}
\label{tab:recommendation_performance}
\end{table}

To verify PRAG's ability for recommendation, we report RMSE scores on the three datasets as shown in Table~\ref{tab:recommendation_performance}.
Since SUM and \optimus~are based on the same retriever model as PRAG, we report PRAG's performance only in this section.
Further, Att2Seq cannot produce rating estimations, and thus we omit the model in the table.
We validate that the performance of baseline models is consistent with previous works on the same dataset~\cite{geng-etal-2022-improving}.
As shown in Table~\ref{tab:recommendation_performance},
\ours~achieved state-of-the-art rating estimation performance.

\subsection{Human Evaluation}

\begin{table}[t!]
\small
\centering
\setlength\tabcolsep{4pt}
\resizebox{0.85\columnwidth}{!}{%
\begin{tabular}{l | cc cc cc c cc cc cc c}
\toprule
\multirow{2}{*}
 & \multicolumn{3}{c}{\textbf{Fluency}}
  &
  & \multicolumn{3}{c}{\textbf{Informativeness}} \\
\cmidrule(lr){2-5}
\cmidrule(lr){6-9}
  \textbf{\ours}~vs.
  & 
  \multicolumn{1}{c}{\textbf{Movie}} & \multicolumn{1}{c}{\textbf{Trip}} & \multicolumn{1}{c}{\textbf{Yelp}} &
  &
  \multicolumn{1}{c}{\textbf{Movie}} & \multicolumn{1}{c}{\textbf{Trip}} & \multicolumn{1}{c}{\textbf{Yelp}} &
\\
\midrule
\nrt
               & \textbf{37}
               & \textbf{16}
               & \textbf{13}
               &
               & \textbf{28}
               & \textbf{1}        
               & \textbf{3} 
\\
\peter
               & \textbf{20}
               & \textbf{26}
               & \textbf{15}
               &
               & -2
               & \textbf{12}       
               & \textbf{27} 
\\
SUM
               & \textbf{8}
               & \textbf{5}
               & \textbf{3}
               &
               & 0
               & -6  
               & \textbf{5} 
\\
    \bottomrule
    \end{tabular}%
}
    \caption{\textbf{Human evaluation results} of \ours~versus baseline models. The number reported are the number of \textit{additional} times \ours~win against the opponent model. Note both \ours~and SUM are retriever augmented generators.} 
\label{tab:human1}
\end{table}

We perform human evaluation using 150 test samples with several of the strongest baselines. 
Specifically, we compare \ours's performance against \nrt, \peter~and SUM. 
We omit \atttoseq~and \pepler~since the models are not as performant as other models in recommendation performance, and leave \peter+ out since ground-truth aspects are not always available. 
Finally, we pick the summarizer as a stronger retriever-augmented baseline (compared to \optimus) based on automatic evaluation results. 
We compare the generated reviews in terms of (1) \textbf{fluency} compared to other models (2) \textbf{informativeness} to the user. 
We report the results in Table~\ref{tab:human1}. 
Similar to automatic evaluation results, the retriever could reliably boost generation performance: \ours~is consistently ranked as the most fluent model by human evaluators, while SUM has almost on-par performance due to having access to the personalized retriever.
Meanwhile, \ours~ and SUM almost invariably win in terms of informativeness compared to previous works.

\section{Analysis and Discussion}

\label{sec:verifying_decision_choices}

\paragraph{Analysis of marginalization and retrieval reliability.}

To validate whether marginalizing the predicted embedding before performing retrieval truly uncovers user or item characteristics instead of resulting in random vectors, we perform a simple test by comparing the retrieval result between a pair of mirroring retrievers. 
Specifically, we hypothesize that given a user and an item, a retriever that ties $Q_{u,i}$ to $\gamma_u$ and a retriever that ties $Q_{u,i}$ to $\gamma_i$ should achieve agreement after marginalization.
That is, the two retrievers should be able to retrieve similar reviews.
To verify this, we report the average exact-match between two retrievers using agreement at 5.
As shown in Table~\ref{tab:cross_retriever_agreement}, the retrievers consistently achieved significantly higher agreement compared to a random baseline across three datasets.
Further, we report examples of retrieved reviews with and without marginalization.
As shown in Table~\ref{tab:qualitative_analysis_of_marginalization}, after marginalization, there is a clear trend that reviews related to aesthetic aspects are being retrieved, as opposed to a set of generic reviews without marginalization.

\paragraph{Analysis of sentence embedding model.}

We hypothesize that any strong sentence embedding model could be used for PRAG. 
To validate such a hypothesis, we train \ours~model on all three datasets using a pre-trained T5-based sentence encoder. 
We report the performance as in Table~\ref{tab:recommendation_performance}. 
As shown, there is no significant performance variation across different types of sentence embeddings. 

\begin{table}[t!]
\small
\setlength\tabcolsep{11pt}
\center
\begin{tabular}{l}
\toprule
\bf Retrieved Reviews \\
\midrule
\begin{minipage}{75mm}
\underline{the decor} of the hotel is greatly refined. ...
the building and \underline{decoration} are very nice and very \underline{tastefully decorated}. 
a four seasons in every respect it is an \underline{architecturally interesting and esthetically pleasing} property in a great location. 
the rooms are \underline{stunning}. %
\end{minipage}
\\
\bottomrule
\end{tabular}
\caption{\textbf{Retrieved reviews} (cropped, from top-5 results). With marginalization, the retrieved reviews consistently focus a \underline{specific aspect}. See appendix for comparison with non-marginalized retrieval results.
}
\label{tab:qualitative_analysis_of_marginalization}
\end{table}

\begin{table}[t!]
\small
\setlength\tabcolsep{11pt}
\center
\begin{tabular}{lccccc}
\toprule
{} &  \bf Movie & \bf TripAdivor &  \bf Yelp  \\
\midrule
Retriever  &  2.87 & 2.44 &      2.50  \\
Random  & 0.54 & 0.61 &        0.27  \\
\bottomrule
\end{tabular}
\caption{\textbf{Average agreement-at-5} of mirroring retrievers. 
}
\label{tab:cross_retriever_agreement}
\end{table}

\paragraph{Effect of tying user or item Factors to latent query.}

Similar to the HFT~\cite{mcauley2013hidden} model, the rating estimation component in \ours~could either tie the user or item factor to $Q_{u,i}$. 
We conduct experiments using both types of architectures, and also report our results in Table~\ref{tab:recommendation_performance}. 
Similar to findings reported for the HFT model~\cite{mcauley2013hidden}, the performance is generally dataset-dependent, and the design could be viewed as a hyper-parameter.

\section{Summary and Outlook}

In this work, we propose PRAG, a retriever-reader model that can generate factual and diverse explanations for recommendation. 
Experiments on three real-world datasets show \ours~can generate both factually grounded and informative explanations. 
We also investigate the cause of hallucinated content in review generation, and demonstrate the benefit of training text generation models on hallucination-free tasks and datasets. 
Meanwhile, although we adopted a question-answering model for explanation generation, PRAG's retrieval component could provide support for personalizing a wider range of knowledge-based tasks, such as personalized conversational recommendation, summarization, and product description generation.

\bibliography{arxiv23}

\begin{thebibliography}{44}
\providecommand{\natexlab}[1]{#1}

\bibitem[{Amplayo and Lapata(2020)}]{amplayo-lapata-2020-unsupervised}
Amplayo, R.~K.; and Lapata, M. 2020.
\newblock Unsupervised Opinion Summarization with Noising and Denoising.
\newblock In \emph{ACL}, 1934--1945. Online: Association for Computational
  Linguistics.

\bibitem[{Bra{\v{z}}inskas, Lapata, and
  Titov(2020)}]{brazinskas-etal-2020-unsupervised}
Bra{\v{z}}inskas, A.; Lapata, M.; and Titov, I. 2020.
\newblock Unsupervised Opinion Summarization as Copycat-Review Generation.
\newblock In \emph{ACL}, 5151--5169. Online: Association for Computational
  Linguistics.

\bibitem[{Bra{\v{z}}inskas, Lapata, and
  Titov(2021)}]{brazinskas-etal-2021-learning}
Bra{\v{z}}inskas, A.; Lapata, M.; and Titov, I. 2021.
\newblock Learning Opinion Summarizers by Selecting Informative Reviews.
\newblock In \emph{EMNLP}, 9424--9442. Online and Punta Cana, Dominican
  Republic: Association for Computational Linguistics.

\bibitem[{Brown et~al.(2020)Brown, Mann, Ryder, Subbiah, Kaplan, Dhariwal,
  Neelakantan, Shyam, Sastry, Askell, Agarwal, Herbert-Voss, Krueger, Henighan,
  Child, Ramesh, Ziegler, Wu, Winter, Hesse, Chen, Sigler, Litwin, Gray, Chess,
  Clark, Berner, McCandlish, Radford, Sutskever, and
  Amodei}]{Brown2020LanguageMA}
Brown, T.~B.; Mann, B.; Ryder, N.; Subbiah, M.; Kaplan, J.; Dhariwal, P.;
  Neelakantan, A.; Shyam, P.; Sastry, G.; Askell, A.; Agarwal, S.;
  Herbert-Voss, A.; Krueger, G.; Henighan, T.~J.; Child, R.; Ramesh, A.;
  Ziegler, D.~M.; Wu, J.; Winter, C.; Hesse, C.; Chen, M.; Sigler, E.; Litwin,
  M.; Gray, S.; Chess, B.; Clark, J.; Berner, C.; McCandlish, S.; Radford, A.;
  Sutskever, I.; and Amodei, D. 2020.
\newblock Language Models are Few-Shot Learners.
\newblock \emph{ArXiv}, abs/2005.14165.

\bibitem[{Cheng et~al.(2016)Cheng, Koc, Harmsen, Shaked, Chandra, Aradhye,
  Anderson, Corrado, Chai, Ispir, Anil, Haque, Hong, Jain, Liu, and
  Shah}]{cheng2016widedeep}
Cheng, H.-T.; Koc, L.; Harmsen, J.; Shaked, T.; Chandra, T.; Aradhye, H.;
  Anderson, G.; Corrado, G.; Chai, W.; Ispir, M.; Anil, R.; Haque, Z.; Hong,
  L.; Jain, V.; Liu, X.; and Shah, H. 2016.
\newblock Wide \& Deep Learning for Recommender Systems.
\newblock In \emph{1st Workshop on Deep Learning for Recommender Systems}, DLRS
  2016, 7–10. New York, NY, USA: Association for Computing Machinery.
\newblock ISBN 9781450347952.

\bibitem[{Dalvi et~al.(2021)Dalvi, Jansen, Tafjord, Xie, Smith, Pipatanangkura,
  and Clark}]{dalvi-etal-2021-explaining}
Dalvi, B.; Jansen, P.; Tafjord, O.; Xie, Z.; Smith, H.; Pipatanangkura, L.; and
  Clark, P. 2021.
\newblock Explaining Answers with Entailment Trees.
\newblock In \emph{EMNLP}, 7358--7370. Online and Punta Cana, Dominican
  Republic: Association for Computational Linguistics.

\bibitem[{Devlin et~al.(2019)Devlin, Chang, Lee, and
  Toutanova}]{devlin-etal-2019-bert}
Devlin, J.; Chang, M.-W.; Lee, K.; and Toutanova, K. 2019.
\newblock {BERT}: Pre-training of Deep Bidirectional Transformers for Language
  Understanding.
\newblock In \emph{NAACL-HLT}, 4171--4186. Minneapolis, Minnesota: Association
  for Computational Linguistics.

\bibitem[{Dong et~al.(2017)Dong, Huang, Wei, Lapata, Zhou, and
  Xu}]{dong-etal-2017-learning-generate}
Dong, L.; Huang, S.; Wei, F.; Lapata, M.; Zhou, M.; and Xu, K. 2017.
\newblock Learning to Generate Product Reviews from Attributes.
\newblock In \emph{EACL}, 623--632. Valencia, Spain: Association for
  Computational Linguistics.

\bibitem[{Geng et~al.(2022)Geng, Fu, Ge, Li, de~Melo, and
  Zhang}]{geng-etal-2022-improving}
Geng, S.; Fu, Z.; Ge, Y.; Li, L.; de~Melo, G.; and Zhang, Y. 2022.
\newblock Improving Personalized Explanation Generation through Visualization.
\newblock In \emph{ACL}, 244--255. Dublin, Ireland: Association for
  Computational Linguistics.

\bibitem[{Guu et~al.(2020)Guu, Lee, Tung, Pasupat, and Chang}]{guu2020realm}
Guu, K.; Lee, K.; Tung, Z.; Pasupat, P.; and Chang, M.-W. 2020.
\newblock {REALM}: Retrieval-augmented language model pre-training.
\newblock \emph{arXiv preprint arXiv:2002.08909}.

\bibitem[{Hada, Vijaikumar, and Shevade(2021)}]{Hada2021ReXPlugER}
Hada, D.~V.; Vijaikumar, M.; and Shevade, S.~K. 2021.
\newblock ReXPlug: Explainable Recommendation using Plug-and-Play Language
  Model.
\newblock \emph{SIGIR}.

\bibitem[{Hancock et~al.(2018)Hancock, Varma, Wang, Bringmann, Liang, and
  R{\'e}}]{hancock-etal-2018-training}
Hancock, B.; Varma, P.; Wang, S.; Bringmann, M.; Liang, P.; and R{\'e}, C.
  2018.
\newblock Training Classifiers with Natural Language Explanations.
\newblock In \emph{ACL}, 1884--1895. Melbourne, Australia: Association for
  Computational Linguistics.

\bibitem[{He and McAuley(2016)}]{He2016UpsAD}
He, R.; and McAuley, J. 2016.
\newblock Ups and Downs: Modeling the Visual Evolution of Fashion Trends with
  One-Class Collaborative Filtering.
\newblock \emph{WWW}.

\bibitem[{Hokamp and Liu(2017)}]{hokamp-liu-2017-lexically}
Hokamp, C.; and Liu, Q. 2017.
\newblock Lexically Constrained Decoding for Sequence Generation Using Grid
  Beam Search.
\newblock In \emph{ACL}, 1535--1546. Vancouver, Canada: Association for
  Computational Linguistics.

\bibitem[{Iso et~al.(2021)Iso, Wang, Suhara, Angelidis, and
  Tan}]{iso-etal-2021-convex-aggregation}
Iso, H.; Wang, X.; Suhara, Y.; Angelidis, S.; and Tan, W.-C. 2021.
\newblock {C}onvex {A}ggregation for {O}pinion {S}ummarization.
\newblock In \emph{Findings of EMNLP}, 3885--3903. Punta Cana, Dominican
  Republic: Association for Computational Linguistics.

\bibitem[{Jhamtani et~al.(2018)Jhamtani, Gangal, Hovy, Neubig, and
  Berg-Kirkpatrick}]{jhamtani-etal-2018-learning}
Jhamtani, H.; Gangal, V.; Hovy, E.; Neubig, G.; and Berg-Kirkpatrick, T. 2018.
\newblock Learning to Generate Move-by-Move Commentary for Chess Games from
  Large-Scale Social Forum Data.
\newblock In \emph{ACL}, 1661--1671. Melbourne, Australia: Association for
  Computational Linguistics.

\bibitem[{Lewis et~al.(2020)Lewis, Perez, Piktus, Petroni, Karpukhin, Goyal,
  K{\"{u}}ttler, Lewis, Yih, Rockt{\"{a}}schel, Riedel, and
  Kiela}]{DBLP:conf/nips/LewisPPPKGKLYR020}
Lewis, P. S.~H.; Perez, E.; Piktus, A.; Petroni, F.; Karpukhin, V.; Goyal, N.;
  K{\"{u}}ttler, H.; Lewis, M.; Yih, W.; Rockt{\"{a}}schel, T.; Riedel, S.; and
  Kiela, D. 2020.
\newblock Retrieval-Augmented Generation for Knowledge-Intensive {NLP} Tasks.
\newblock In Larochelle, H.; Ranzato, M.; Hadsell, R.; Balcan, M.; and Lin, H.,
  eds., \emph{NeurIPS}.

\bibitem[{Li et~al.(2020)Li, Gao, Li, Peng, Li, Zhang, and
  Gao}]{li-etal-2020-optimus}
Li, C.; Gao, X.; Li, Y.; Peng, B.; Li, X.; Zhang, Y.; and Gao, J. 2020.
\newblock Optimus: Organizing Sentences via Pre-trained Modeling of a Latent
  Space.
\newblock In \emph{EMNLP}, 4678--4699. Online: Association for Computational
  Linguistics.

\bibitem[{Li et~al.(2016)Li, Galley, Brockett, Gao, and
  Dolan}]{li-etal-2016-diversity}
Li, J.; Galley, M.; Brockett, C.; Gao, J.; and Dolan, B. 2016.
\newblock A Diversity-Promoting Objective Function for Neural Conversation
  Models.
\newblock In \emph{NAACL-HLT}, 110--119. San Diego, California: Association for
  Computational Linguistics.

\bibitem[{Li, Zhang, and Chen(2020)}]{li_2020_generate_neural_template}
Li, L.; Zhang, Y.; and Chen, L. 2020.
\newblock Generate Neural Template Explanations for Recommendation.
\newblock In \emph{CIKM}, CIKM '20, 755–764. New York, NY, USA: Association
  for Computing Machinery.
\newblock ISBN 9781450368599.

\bibitem[{Li, Zhang, and Chen(2021)}]{li-etal-2021-personalized}
Li, L.; Zhang, Y.; and Chen, L. 2021.
\newblock Personalized Transformer for Explainable Recommendation.
\newblock In \emph{ACL-IJCNLP}, 4947--4957. Online: Association for
  Computational Linguistics.

\bibitem[{Li, Zhang, and Chen(2022)}]{2022-PEPLER}
Li, L.; Zhang, Y.; and Chen, L. 2022.
\newblock Personalized Prompt Learning for Explainable Recommendation.
\newblock \emph{arXiv preprint arXiv:2202.07371}.

\bibitem[{Li et~al.(2017)Li, Wang, Ren, Bing, and Lam}]{Li2017NeuralRR}
Li, P.; Wang, Z.; Ren, Z.; Bing, L.; and Lam, W. 2017.
\newblock Neural Rating Regression with Abstractive Tips Generation for
  Recommendation.
\newblock \emph{SIGIR}.

\bibitem[{Liu et~al.(2022)Liu, Sabour, Zheng, Ke, Zhu, and
  Huang}]{DBLP:conf/acl/LiuSZK0H22}
Liu, S.; Sabour, S.; Zheng, Y.; Ke, P.; Zhu, X.; and Huang, M. 2022.
\newblock Rethinking and Refining the Distinct Metric.
\newblock In Muresan, S.; Nakov, P.; and Villavicencio, A., eds., \emph{ACL},
  762--770. Association for Computational Linguistics.

\bibitem[{Longpre et~al.(2021)Longpre, Perisetla, Chen, Ramesh, DuBois, and
  Singh}]{longpre-etal-2021-entity}
Longpre, S.; Perisetla, K.; Chen, A.; Ramesh, N.; DuBois, C.; and Singh, S.
  2021.
\newblock Entity-Based Knowledge Conflicts in Question Answering.
\newblock In \emph{EMNLP}, 7052--7063. Online and Punta Cana, Dominican
  Republic: Association for Computational Linguistics.

\bibitem[{Majumder et~al.(2021)Majumder, Berg-Kirkpatrick, McAuley, and
  Jhamtani}]{majumder-etal-2021-unsupervised}
Majumder, B.~P.; Berg-Kirkpatrick, T.; McAuley, J.; and Jhamtani, H. 2021.
\newblock Unsupervised Enrichment of Persona-grounded Dialog with Background
  Stories.
\newblock In \emph{ACL-IJCNLP}, 585--592. Online: Association for Computational
  Linguistics.

\bibitem[{Majumder et~al.(2022)Majumder, Camburu, Lukasiewicz, and
  McAuley}]{DBLP:journals/corr/abs-2106-13876}
Majumder, B.~P.; Camburu, O.; Lukasiewicz, T.; and McAuley, J.~J. 2022.
\newblock Rationale-Inspired Natural Language Explanations with Commonsense.
\newblock \emph{ICML}, abs/2106.13876.

\bibitem[{Marasovi{\'c} et~al.(2020)Marasovi{\'c}, Bhagavatula, Park, Le~Bras,
  Smith, and Choi}]{marasovic-etal-2020-natural}
Marasovi{\'c}, A.; Bhagavatula, C.; Park, J.~s.; Le~Bras, R.; Smith, N.~A.; and
  Choi, Y. 2020.
\newblock Natural Language Rationales with Full-Stack Visual Reasoning: From
  Pixels to Semantic Frames to Commonsense Graphs.
\newblock In \emph{Findings of EMNLP}, 2810--2829. Online: Association for
  Computational Linguistics.

\bibitem[{Maynez et~al.(2020)Maynez, Narayan, Bohnet, and
  McDonald}]{maynez-etal-2020-faithfulness}
Maynez, J.; Narayan, S.; Bohnet, B.; and McDonald, R. 2020.
\newblock On Faithfulness and Factuality in Abstractive Summarization.
\newblock In \emph{ACL}, 1906--1919. Online: Association for Computational
  Linguistics.

\bibitem[{McAuley and Leskovec(2013)}]{mcauley2013hidden}
McAuley, J.; and Leskovec, J. 2013.
\newblock Hidden Factors and Hidden Topics: Understanding Rating Dimensions
  with Review Text.
\newblock In \emph{ACM RecSys}, RecSys '13, 165–172. New York, NY, USA:
  Association for Computing Machinery.
\newblock ISBN 9781450324090.

\bibitem[{Ni et~al.(2017)Ni, Lipton, Vikram, and
  McAuley}]{ni-etal-2017-estimating}
Ni, J.; Lipton, Z.~C.; Vikram, S.; and McAuley, J. 2017.
\newblock Estimating Reactions and Recommending Products with Generative Models
  of Reviews.
\newblock In \emph{IJCNLP}, 783--791. Taipei, Taiwan: Asian Federation of
  Natural Language Processing.

\bibitem[{Ni and McAuley(2018)}]{ni-mcauley-2018-personalized}
Ni, J.; and McAuley, J. 2018.
\newblock Personalized Review Generation By Expanding Phrases and Attending on
  Aspect-Aware Representations.
\newblock In \emph{ACL}, 706--711. Melbourne, Australia: Association for
  Computational Linguistics.

\bibitem[{Pagnoni, Balachandran, and
  Tsvetkov(2021)}]{pagnoni-etal-2021-understanding}
Pagnoni, A.; Balachandran, V.; and Tsvetkov, Y. 2021.
\newblock Understanding Factuality in Abstractive Summarization with {FRANK}: A
  Benchmark for Factuality Metrics.
\newblock In \emph{NAACL-HLT}, 4812--4829. Online: Association for
  Computational Linguistics.

\bibitem[{Pillutla et~al.(2021)Pillutla, Swayamdipta, Zellers, Thickstun,
  Welleck, Choi, and Harchaoui}]{pillutla-etal:mauve:neurips2021}
Pillutla, K.; Swayamdipta, S.; Zellers, R.; Thickstun, J.; Welleck, S.; Choi,
  Y.; and Harchaoui, Z. 2021.
\newblock MAUVE: Measuring the Gap Between Neural Text and Human Text using
  Divergence Frontiers.
\newblock In \emph{NeurIPS}.

\bibitem[{Raffel et~al.(2020)Raffel, Shazeer, Roberts, Lee, Narang, Matena,
  Zhou, Li, and Liu}]{DBLP:journals/jmlr/RaffelSRLNMZLL20}
Raffel, C.; Shazeer, N.; Roberts, A.; Lee, K.; Narang, S.; Matena, M.; Zhou,
  Y.; Li, W.; and Liu, P.~J. 2020.
\newblock Exploring the Limits of Transfer Learning with a Unified Text-to-Text
  Transformer.
\newblock \emph{J. Mach. Learn. Res.}, 21: 140:1--140:67.

\bibitem[{Ram et~al.(2021)Ram, Kirstain, Berant, Globerson, and
  Levy}]{ram-etal-2021-shot}
Ram, O.; Kirstain, Y.; Berant, J.; Globerson, A.; and Levy, O. 2021.
\newblock Few-Shot Question Answering by Pretraining Span Selection.
\newblock In \emph{ACL-IJCNLP}, 3066--3079. Online: Association for
  Computational Linguistics.

\bibitem[{Song et~al.(2020)Song, Tan, Qin, Lu, and
  Liu}]{DBLP:conf/nips/Song0QLL20}
Song, K.; Tan, X.; Qin, T.; Lu, J.; and Liu, T. 2020.
\newblock MPNet: Masked and Permuted Pre-training for Language Understanding.
\newblock In Larochelle, H.; Ranzato, M.; Hadsell, R.; Balcan, M.; and Lin, H.,
  eds., \emph{NeurIPS}.

\bibitem[{Tafjord and Clark(2021)}]{DBLP:journals/corr/macaw}
Tafjord, O.; and Clark, P. 2021.
\newblock General-Purpose Question-Answering with Macaw.
\newblock \emph{CoRR}, abs/2109.02593.

\bibitem[{Wiegreffe, Marasovi{\'c}, and
  Smith(2021)}]{wiegreffe-etal-2021-measuring}
Wiegreffe, S.; Marasovi{\'c}, A.; and Smith, N.~A. 2021.
\newblock {M}easuring Association Between Labels and Free-Text Rationales.
\newblock In \emph{EMNLP}, 10266--10284. Online and Punta Cana, Dominican
  Republic: Association for Computational Linguistics.

\bibitem[{Wolf et~al.(2020)Wolf, Debut, Sanh, Chaumond, Delangue, Moi, Cistac,
  Rault, Louf, Funtowicz, Davison, Shleifer, von Platen, Ma, Jernite, Plu, Xu,
  Le~Scao, Gugger, Drame, Lhoest, and Rush}]{wolf-etal-2020-transformers}
Wolf, T.; Debut, L.; Sanh, V.; Chaumond, J.; Delangue, C.; Moi, A.; Cistac, P.;
  Rault, T.; Louf, R.; Funtowicz, M.; Davison, J.; Shleifer, S.; von Platen,
  P.; Ma, C.; Jernite, Y.; Plu, J.; Xu, C.; Le~Scao, T.; Gugger, S.; Drame, M.;
  Lhoest, Q.; and Rush, A. 2020.
\newblock Transformers: State-of-the-Art Natural Language Processing.
\newblock In \emph{EMNLP: System Demonstrations}, 38--45. Online: Association
  for Computational Linguistics.

\bibitem[{Zhang et~al.(2014)Zhang, Lai, Zhang, Zhang, Liu, and
  Ma}]{DBLP:conf/sigir/ZhangL0ZLM14}
Zhang, Y.; Lai, G.; Zhang, M.; Zhang, Y.; Liu, Y.; and Ma, S. 2014.
\newblock Explicit factor models for explainable recommendation based on
  phrase-level sentiment analysis.
\newblock In Geva, S.; Trotman, A.; Bruza, P.; Clarke, C. L.~A.; and
  J{\"{a}}rvelin, K., eds., \emph{SIGIR}, 83--92. {ACM}.

\bibitem[{Zhou et~al.(2018)Zhou, Zhu, Song, Fan, Zhu, Ma, Yan, Jin, Li, and
  Gai}]{Zhou2018DIN}
Zhou, G.; Zhu, X.; Song, C.; Fan, Y.; Zhu, H.; Ma, X.; Yan, Y.; Jin, J.; Li,
  H.; and Gai, K. 2018.
\newblock Deep Interest Network for Click-Through Rate Prediction.
\newblock In \emph{SIGKDD}, KDD '18, 1059–1068. New York, NY, USA:
  Association for Computing Machinery.
\newblock ISBN 9781450355520.

\bibitem[{Zhou et~al.(2020)Zhou, Hu, Zhang, Liang, Sun, Xiong, and
  Tang}]{DBLP:conf/nips/ZhouHZLSXT20}
Zhou, W.; Hu, J.; Zhang, H.; Liang, X.; Sun, M.; Xiong, C.; and Tang, J. 2020.
\newblock Towards Interpretable Natural Language Understanding with
  Explanations as Latent Variables.
\newblock In Larochelle, H.; Ranzato, M.; Hadsell, R.; Balcan, M.; and Lin, H.,
  eds., \emph{NeurIPS}.

\bibitem[{Zhou, Li, and Li(2021)}]{zhou-etal-2021-learning}
Zhou, W.; Li, Q.; and Li, C. 2021.
\newblock Learning from Perturbations: Diverse and Informative Dialogue
  Generation with Inverse Adversarial Training.
\newblock In \emph{ACL-IJCNLP}, 694--703. Online: Association for Computational
  Linguistics.

\end{thebibliography}

\end{document}